\pdfoutput=1

\documentclass[11pt]{article}

\usepackage{emnlp2021}

\usepackage{times}
\usepackage{latexsym}

\usepackage[T1]{fontenc}

\usepackage[utf8]{inputenc}

\usepackage{microtype}
\usepackage{graphicx}
\usepackage{subfigure}
\usepackage{xfrac}
\usepackage{colortbl}
\usepackage{booktabs}  
\usepackage{amsfonts}  
\usepackage{nicefrac}
\usepackage{latexsym}
\usepackage{multirow}
\usepackage{amsmath}
\usepackage{tabularx}
\usepackage{algorithm}
\usepackage{algpseudocode}
\usepackage{todonotes}

\title{Broaden the Vision: Geo-Diverse Visual Commonsense Reasoning}

\author{
	Da Yin 
\quad	Liunian Harold Li
\quad	Ziniu Hu 
\quad	Nanyun Peng
\quad	Kai-Wei Chang \\
	Computer Science Department, University of California, Los Angeles\\
	{\tt \{da.yin,liunian.harold.li,bull,violetpeng,kwchang\}@cs.ucla.edu}
	\\
	{\tt \href{gd-vcr.github.io}{\textbf{\textcolor{orange}{gd-vcr.github.io}}}}
	\\
}

\begin{document}
\maketitle
\begin{abstract}
Commonsense is defined as the knowledge that is shared by everyone. 
However, certain types of commonsense knowledge are correlated with culture and geographic locations and they are only shared locally. For example, the scenarios of wedding ceremonies vary across regions due to different customs influenced by historical and religious factors. Such regional characteristics, however, are generally omitted in prior work. 
In this paper, we construct a \textbf{G}eo-\textbf{D}iverse \textbf{V}isual \textbf{C}ommonsense \textbf{R}easoning dataset (\textbf{GD-VCR}) to test vision-and-language models' ability to understand cultural and geo-location-specific commonsense. 
In particular, we study two state-of-the-art Vision-and-Language models, VisualBERT and ViLBERT trained on VCR, a standard multimodal commonsense benchmark with images primarily from Western regions. We then evaluate how well the trained models can generalize to answering the questions in \textbf{GD-VCR}. We find that the performance of both models for non-Western regions including East Asia, South Asia, and Africa is significantly lower than that for Western region.
We analyze the reasons behind the performance disparity and find that the performance gap is larger on QA pairs that: 1) are concerned with culture-related scenarios, e.g., weddings, religious activities, and festivals; 2) require high-level geo-diverse commonsense reasoning rather than low-order perception and recognition. Dataset and code are released at \url{https://github.com/WadeYin9712/GD-VCR}.
\end{abstract}

\section{Introduction}
Commonsense reasoning endows machines with high-level reasoning ability to understand situations with implicit commonsense knowledge. 
Suppose that there is a scene where a woman is wearing a bridal gown at a party. An ideal AI system with commonsense knowledge should be able to infer that this woman is attending a wedding and likely to be the bride.

Recently, the field of commonsense reasoning is progressing with the development of large-scale benchmark datasets~\cite{zellers-etal-2018-swag,talmor-etal-2019-commonsenseqa}, intended to cover a wide range of commonsense knowledge, such as physical interactions \cite{Bisk2020}, social conventions \cite{sap-etal-2019-social}, and commonsense grounded in vision \cite{zellers2019vcr}.

\begin{figure*}[t]
    \centering
    \includegraphics[width=0.95\linewidth, trim=0 240 75 0, clip]{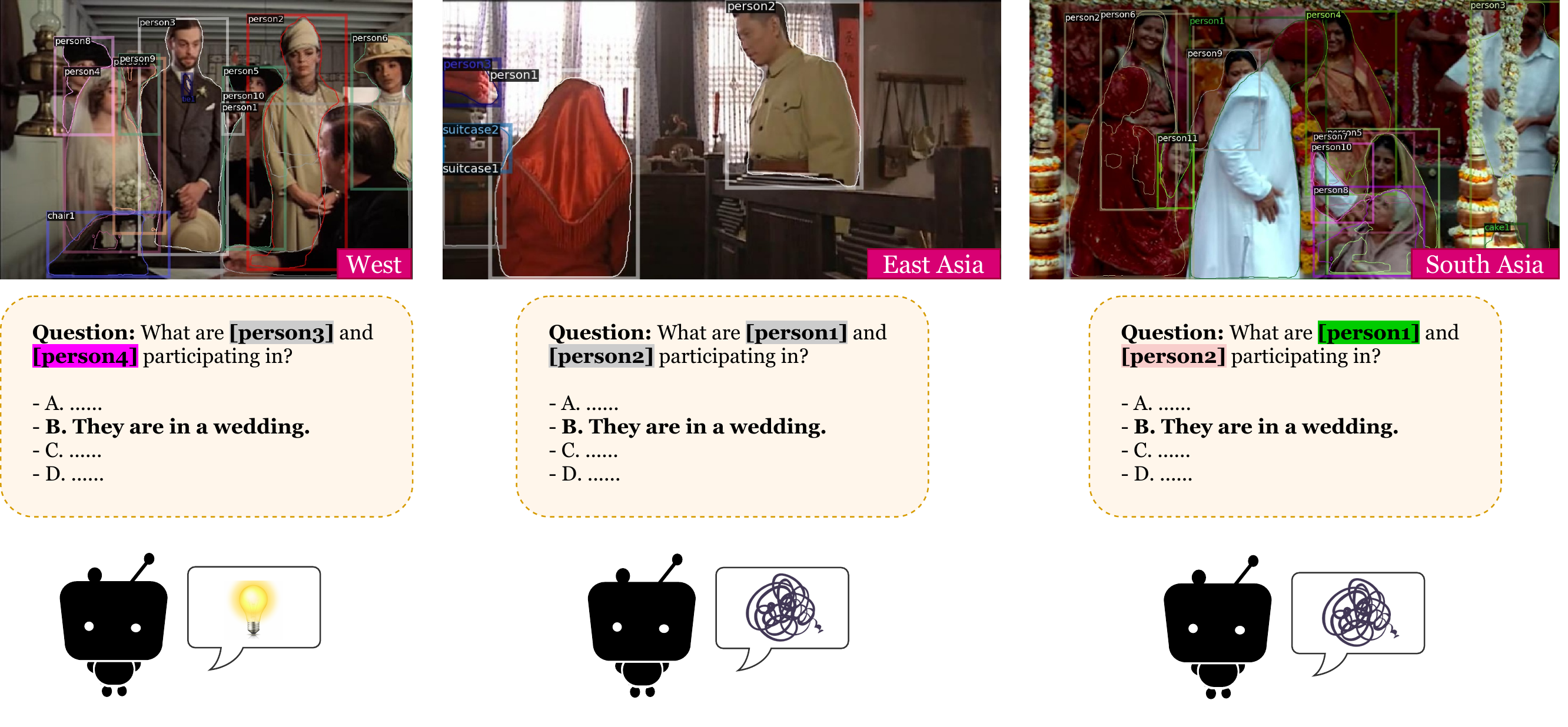}
    \caption{Examples in GD-VCR. The three images are all about weddings but from different regions (left-to-right order: Western, East Asian, South Asian). Current Vision-and-Language models perform well on answering questions about Western weddings but often make mistakes when encountering wedding scenarios in other regions.}
    \label{fig:intro}
\end{figure*}

However, existing benchmarks are often composed by data from sources in certain regions (e.g., Western movies) and overlook the differences across groups in different regions\footnote{Due to resource constraints, we use regions as a proxy to evaluate commonsense among different groups. We note that groups of individuals in the same region may have different beliefs, cultures, and behaviors. Please see discussion in the section of Ethical Considerations.} due to factors including cultural differences.
In the aforementioned wedding example, while brides are usually in white in Western weddings, they often wear red and their faces are covered with a red cloth in traditional Chinese weddings. 
If a model is unaware of regional characteristics or incapable of capturing the nuanced regional characteristics, it leads to a disparity in performance across different regions~\cite{Acharya2020AnAO}. 
This motivates us to study how well a model trained on existing commonsense annotations can generalize to tasks requiring commonsense beyond Western regions.

In this paper, we mainly focus on regional commonsense with \emph{visual scenes}.
As shown in Figure~\ref{fig:intro}, the three images all describe a wedding but the dresses of the grooms and brides are different, reflecting the regional characteristics of the wedding scenario. 
In this paper, we introduce a new \emph{evaluation} benchmark, \textbf{G}eo-\textbf{D}iverse \textbf{V}isual \textbf{C}ommonsense \textbf{R}easoning (\textbf{GD-VCR}), following the settings of the visual commonsense reasoning (VCR) task~\cite{zellers2019vcr}.
VCR consists of multiple-choice questions paired with images extracted from movies or TV series primarily in \emph{Western} regions. 
GD-VCR includes 328 images, which are mainly sourced from movies and TV series in East Asian, South Asian, African, and Western countries. The images are paired with 886 QA pairs, which need to be answered with geo-diverse commonsense and thorough understanding of the images.
An example is given in Figure~\ref{fig:intro}. GD-VCR benchmark addresses geo-diverse commonsense, such as ``\textit{What are \colorbox[rgb]{ .891,  .891,  .891}{\textbf{[person1]}} and \colorbox[rgb]{ .891,  .891,  .891}{\textbf{[person2]}} participating?}''. With the help of these questions, it can manifest how models behave differently and reveal potential issues with geographical bias in commonsense reasoning. 
GD-VCR is one of the first benchmarks to evaluate model's reasoning ability on the task which requires geo-specific commonsense knowledge.

We first study \emph{how well a model trained on VCR can generalize to questions involving geo-specific commonsense}. Experimenting with two pre-trained vision-and-language (V\&L) models, VisualBERT~\cite{li2019visualbert} and ViLBERT~\cite{Lu2019ViLBERTPT}, we observe that two models achieve 64\%-68\% accuracy over the QA pairs on images from Western regions, while their accuracy on images from East Asian region ranges around 45\%-51\%. The significant performance disparity suggests that the commonsense learned in these models cannot be generalized well across different regions.

We further investigate \emph{the reasons why the model exhibits such disparity} based on the results of VisualBERT. We first find that the performance gap on the images from Western and non-Western regions is large on the scenarios involving regional characteristics, such as weddings, religion and festivals.
We also discover that the disparity is related to the reasoning difficulty of QA pairs. On the QA pairs only requiring basic visual recognition, e.g., ``\textit{What's \colorbox[rgb]{ .891,  .891,  .891}{\textbf{[person3]}} wearing? \colorbox[rgb]{ .891,  .891,  .891}{\textbf{[person3]}} is wearing a suit.}'', the model achieves relatively similar performance over the four regions; however, the gap enlarges when the questions involve higher-level reasoning with commonsense and rich visual contexts.

By presenting the GD-VCR benchmark, we call upon the researchers to empower AI systems with geo-diverse commonsense such that they are capable of conducting high-level reasoning on data from different regions.

\section{Related Work}
\paragraph{Commonsense Reasoning Benchmarks.}
Recently, there has been an emergence of commonsense reasoning benchmarks~\cite{zellers2019vcr,talmor-etal-2019-commonsenseqa,sap-etal-2019-social,zhou-etal-2019-going,huang-etal-2019-cosmos,bhagavatula2020abductive,Bisk2020}, which cover a great variety of commonsense knowledge including visual, social, physical, and temporal commonsense. 
However, these commonsense benchmarks are mostly constructed by annotators from certain regions (e.g., the US and UK) using specific languages (e.g., English). XCOPA~\cite{ponti-etal-2020-xcopa} and X-CSR~\cite{lin-etal-2021-common} are two multilingual benchmarks, but most samples in both benchmarks are simply translated from English and cannot reflect the regional characteristics. Different from previous benchmarks, GD-VCR focuses on geo-diverse commonsense instead of viewing commonsense as a universal monolith.

\paragraph{Vision-and-Language Tasks.}
A long line of research seeks to build vision-and-language datasets that test a model's ability to understand the visual world and how it is grounded in natural language. The tasks take on various forms, such as phrase grounding~\cite{kazemzadeh-etal-2014-referitgame,plummer2015flickr30k}, visual question answering~\cite{antol2015vqa,balanced_vqa_v2}, and visual reasoning~\cite{zellers2019vcr,suhr-etal-2019-corpus}.
To solve these tasks, a wide range of visual grounding skills are required.  
However, in existing tasks, little consideration is taken into reasoning on the images with regional characteristics. 

\paragraph{Geographic Bias.}
Geographic bias is a serious issue that may cause harmful effects on certain groups of people. In computer vision, researchers~\cite{shankar2017no,de2019does} find that most images from two large-scale image datasets, ImageNet~\cite{imagenet_cvpr09} and OpenImages~\cite{openimages}, are amerocentric and eurocentric. When a model trained on these datasets is applied to images from other regions, the performance will drop drastically. There also exists geographic bias in language technology. For example, it underlies natural language processing~\cite{blodgett-etal-2016-demographic,jurgens-etal-2017-incorporating,ghosh2021cross} and automatic speech recognition~\cite{tatman-2017-gender,koenecke2020racial} models. Our work seeks to reveal and test the geographic bias in the visual commonsense reasoning task and models.

\section{Benchmark Construction}
\begin{figure*}[t]
    \centering
    \includegraphics[width=0.9\linewidth, trim=5 45 5 0, clip]{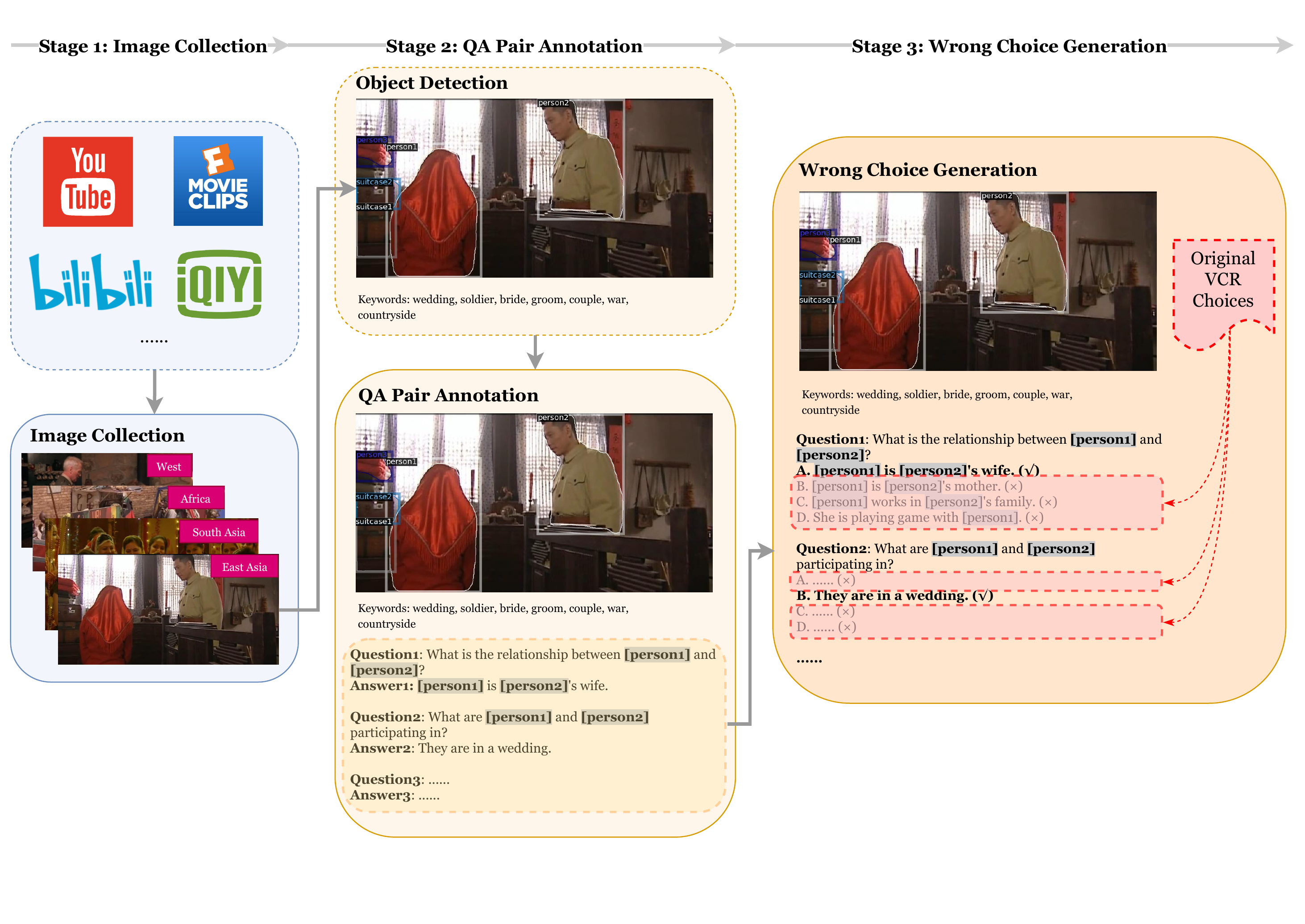}
    \caption{Overall annotation pipeline. It is divided into three stages: Stage 1 is to collect images with regional characteristics; Stage 2 is to design QA pairs based on the detected objects; Stage 3 is to generate answer candidates to complete the dataset with the help of the answer choices in the original VCR development set.}
    \label{fig:annotation}
\end{figure*}

To build a geo-diverse visual commonsense reasoning benchmark, we design a three-stage annotation pipeline, following the original VCR dataset. 1) We first ask annotators to collect images from movies and TV series in Western, East Asian, South Asian, and African countries. 2) We request annotators to design questions and write down the right answers according to the collected images. 3) We generate answer candidates automatically and formulate multiple-choice questions. The overview of our pipeline is illustrated in Figure \ref{fig:annotation}. We elaborate on the three stages in the following.

\subsection{Image Collection}
\label{sec:image_collection}
In the image collection stage, we request annotators to follow two principles:
\paragraph{Images with Regional Characteristics.} In our annotation instruction, we require that the collected images should have representative scenarios containing cultural elements of the annotators' regions. We further recommend annotators choose scenarios that are ubiquitous but have specific characteristics across regions, e.g., wedding, funeral, festival, religion events, etc. 
For the purpose of analysis, during image collection, annotators are required to write down keywords on each of their collected images for categorization. For example, the keywords of the middle image in Figure~\ref{fig:intro} are labeled as ``\textit{wedding, soldier, bride, groom, couple, war, countryside}''.

\paragraph{Sources of Images.} Follow the settings of the original VCR dataset, we ask annotators to select diverse and informative images by taking screenshots from movies, TV series, documentaries, and movie trailers from websites including Youtube\footnote{\url{www.youtube.com/}}, BiliBili\footnote{\url{www.bilibili.com/}}, IQIYI\footnote{\url{www.iqiyi.com/}}, etc. These videos usually include various scenarios and rich contents containing a large amount of actions and human interactions. Note that our collected images from Western regions share the same source\footnote{\url{www.youtube.com/c/MOVIECLIPS/videos}} with those in the original VCR development set.
We use them as a control set in the experiments. 
Details are in Appendix \ref{appendix:image_collection_1}.

\subsection{QA Pair Annotation}
\label{sec:qa_annotation}
We recruit another batch of annotators who are familiar with the culture in one of the four regions to annotate QA pairs upon the collected images in English. The annotation stage is divided into two parts: 1) designing questions according to the image contents; 2) annotating the correct answers of the questions.

Following the pre-processing of the original VCR dataset, we first apply the Mask R-CNN object detector\footnote{\url{github.com/facebookresearch/detectron2}. COCO-pretrained Mask R-CNN.} to mark bounding boxes of objects in each image, and the annotators can use the labels (e.g., person, car, bowl, etc.) to design QA pairs.
\subsubsection{Designing Questions}
\label{sec:question_design}
Annotators are asked to design questions based on the following three instructions.
\paragraph{Usage of the Detected Objects.}
Annotators are requested to choose the named objects in the bounding boxes to construct questions. As shown in Figure~\ref{fig:intro}, annotators can design questions such as ``\textit{What is the relationship between \colorbox[rgb]{ .891,  .891,  .891}{\textbf{[person1]}} and \colorbox[rgb]{ .891,  .891,  .891}{\textbf{[person2]}}?}''. This requirement is aligned with the question design in the original VCR dataset.

\paragraph{High-Order Cognitive Questions.}
Following the original VCR dataset, we ask annotators to design \emph{high-order} cognitive questions which require geo-specific commonsense knowledge and visual understanding to be answered. Take the rightmost image in Figure~\ref{fig:intro} as an example. ``\textit{Why is \colorbox[HTML]{99FF33}{\textbf{[person11]}} so happy?}'' is a \emph{qualified} question because people have to observe the surroundings including \colorbox[HTML]{00CC00}{\textbf{[person1]}} and \colorbox[HTML]{F8CECC}{\textbf{[person2]}}'s wearing and others' facial expression, and conclude that it is a wedding. Moreover, \colorbox[HTML]{00CC00}{\textbf{[person1]}} is wearing a wedding dress and others are celebrating for her. Thus, people can infer that \colorbox[HTML]{99FF33}{\textbf{[person11]}} is happy because it is \colorbox[HTML]{00CC00}{\textbf{[person1]}} and \colorbox[HTML]{F8CECC}{\textbf{[person2]}}'s wedding. Overall, to answer this question, we need to combine the image context and commonsense knowledge, and reason with multiple turns. A \emph{disqualified} example of question is ``\textit{What is \colorbox[rgb]{ .891,  .891,  .891}{\textbf{[person3]}} wearing?}'' in the left image of Figure \ref{fig:intro}. It is defined as a \emph{low-order} cognitive question because it can be directly answered by recognizing that the woman is wearing a suit. This type of question does not need commonsense reasoning based on the context information.

\paragraph{Question Templates.}
Since models trained on the original VCR dataset will be evaluated on GD-VCR dataset, we attempt to eliminate the discrepancy of questions used between the original VCR dev set and GD-VCR to mitigate the effect of different question formats. Hence, we ask annotators to design questions by referring to the templates summarized from the original VCR development set. To generate the templates, we first replace nouns, verbs, adjectives, and adverbs of the questions in VCR development set with their POS tags (e.g., NN, VB, JJ, etc.) labeled by NLTK\footnote{\url{www.nltk.org}}, while keeping question words such as ``\textit{what}'', ``\textit{why}'' and ``\textit{how}'', and auxiliary verbs like ``\textit{is}'', ``\textit{do}'' and ``\textit{have}''. In this way, we remove the terms associated with specific questions, while keeping the general patterns. We then apply K-Means~\citep{macqueen1967some} algorithm\footnote{We concatenate the question words and the POS tags of all the other words (e.g., from ``\textit{What is [person1] doing?}'' to ``\textit{What VBZ NNP VBG?}''). Then we use sentence representations of the converted sentences as real-valued vectors in K-Means. The representations are obtained from Sentence-Transformers~\cite{reimers-2019-sentence-bert} based on RoBERTa-base~\cite{DBLP:journals/corr/abs-1907-11692}.} to the question patterns, and manually summarize 17 templates, e.g., ``\textit{What (n.) is sb. v.+ing sth.?}'', ``\textit{What is sb.'s relationship with sb.?}''. Details of the clustering method and the list of question templates are in Appendix~\ref{appendix:qa_pair_annotation}. 

\subsubsection{Annotating Correct Answers}
Annotators are required to write down \emph{only} the \emph{right} answer for the questions they designed. This is to reduce annotation cost and avoid potential annotation artifacts~\cite{zellers2019vcr}. 
We require that the right answers to the questions should be consistent with the image contents. However, we remind annotators to avoid writing answers that are too specific to the video plots because the answers should be inferred without prior knowledge about the plots. In addition, instead of writing named entities or proper names specific to one region, annotators are required to use common expressions in English. 
These instructions would help us maintain the difficulty of GD-VCR to a proper extent. Finally, we invite 3 annotators from each QA pair's corresponding regions to validate the correctness of each question and its answer. If 2 of them have an agreement to approve a certain QA pair, we keep it in the final dataset. 

\subsection{Answer Candidate Generation}
\label{sec:wrong_choice_generation}
In this stage, for each question, we generate three wrong answer candidates (i.e., wrong answer choices), to construct a 4-way multiple-choice QA example. We follow the answer candidate generation algorithm in VCR~\cite{zellers2019vcr}: 
\paragraph{Answer Candidate Pool.} Instead of generating answer candidates from scratch by language models, we leverage the right choices in the original VCR development set, and treat them as an answer candidate pool. All the answer candidates of GD-VCR are derived from this pool. 

\paragraph{Answer Candidate Selection.} The principles for selecting answer candidates from the pool are two-fold: 1) answer candidates should be relevant to questions; 2) they should be dissimilar with the right choice and other selected answer candidates, so that they would not be the right answer incidentally. 
Details of the candidate selection algorithm are in Appendix~\ref{appendix:wrong_choice}.

\begin{table*}[t]
    \centering
    \scalebox{0.73}{
    \begin{tabular}{lccccccc}
    \toprule
    \textbf{Datasets} & \textbf{\# Images} & \textbf{\# QA Pairs} & \textbf{Avg. Len. of Ques.} & \textbf{Avg. Len. of Ans.} & \textbf{Avg. \# Obj.} & \textbf{Avg. \# Relevant Obj.} & \textbf{OOV Rate} \\
    \midrule
    \textbf{Original VCR} & 9929 & 26534 & 6.77 & 7.67 & 10.34 & 2.39 & 12.70\% \\
    \midrule
    \textbf{GD-VCR} & 328 & 886 & 7.38 & 7.68 & 10.18 & 2.38 & 6.75\% \\
    \midrule
    \textbf{ $\circ$ West} & 100 & 275 & 7.36 & 7.19 & 11.10 & 2.28 & 3.44\% \\
    \textbf{ $\circ$ East Asia} & 101 & 282 & 7.59 & 7.59 & 9.57 & 2.42 & 4.50\% \\
    \textbf{ $\circ$ South Asia} & 87 & 221 & 6.85 & 8.00 & 10.29 & 2.12 & 5.49\% \\
    \textbf{ $\circ$ Africa} & 40 & 108 & 7.98 & 8.54 & 9.29 & 3.03 & 7.34\% \\
    \bottomrule
    \end{tabular}%
    }
    \caption{Statistics of the GD-VCR benchmark. The top half of the table is the overall statistics of GD-VCR and the original VCR development set. The bottom half includes the subsets of each region in GD-VCR.}
  \label{tab:datasets}
\end{table*}

\subsection{Dataset Statistics}
Table~\ref{tab:datasets} summarizes the statistics of the GD-VCR benchmark and the original VCR development set.

\paragraph{Texts.} We observe that the average lengths of questions and answers in GD-VCR are similar to those in the original VCR development set. Aside from the text lengths, we also consider out-of-vocabulary (OOV) rate with respect to the original VCR training set. This indicates how much unseen knowledge (e.g., entities specific to certain region) are involved in GD-VCR.
As shown in Table~\ref{tab:datasets}, we find that the OOV rate of the entire GD-VCR dataset is 6.75\%, while that of the original VCR development set is 12.70\%. This shows that GD-VCR has a similar distribution of the vocabulary with the original VCR dataset and the difficulty of GD-VCR does not come from the vocabulary gap. 

\paragraph{Images.}  The average number of the detected objects 10.18 is similar to that of the original VCR development set. Moreover, since the objects mentioned in questions and answers are directly relevant to the reasoning process on each QA pair, we consider statistics of the average number of the relevant objects. The average number of relevant objects in image in GD-VCR is 2.38, which nearly equals that of the VCR development set.

\section{Model Performance and Human Evaluation over \textbf{GD-VCR}}
We are interested in the following questions: 1) Can a model trained on the original VCR dataset (mostly Western scenarios) generalize well to solve reasoning questions require commonsense specific to other regions? 2) Do humans show similar trend when dealing with questions require regional commonsense that they are not familiar with? 

Our experiments are conducted with two Vision-and-Language models VisualBERT~\cite{li2019visualbert} and ViLBERT~\cite{Lu2019ViLBERTPT}. We fine-tuned the two pre-trained models on the original VCR training set and evaluate them on GD-VCR. All the experimental results are the average of 3 runs with different seeds. Implementation details are listed in Appendix \ref{appendix:implementation}.

\begin{table*}[t]
    \centering
    \scalebox{0.9}{
    \begin{tabular}{lcccccc}
        \toprule
        \multirow{2}{*}{\textbf{Datasets}} & \multirow{2}{*}{\textbf{Human}} &
        \multirow{2}{*}{\textbf{Text-only BERT}} & \multicolumn{2}{c}{\textbf{VisualBERT}} & \multicolumn{2}{c}{\textbf{ViLBERT}} \\
                         &   &   & \textbf{Acc.}       & \textbf{Gap (West)}      & \textbf{Acc.}        & \textbf{Gap (West)}                         \\
        \midrule
        \textbf{Original VCR} & - & 53.8$^*$ & 70.10 & +5.73 & 69.84 & +2.57 \\
        \midrule
        \textbf{GD-VCR} & 88.84 & 35.33 & 53.95 & -10.42 & 59.99  & -7.28 \\
        \midrule
        \textbf{ $\circ$ West} & 91.23 & 37.09 & 64.37 & 0.00 &  67.27  &  0.00  \\
        \textbf{ $\circ$ South Asia} & 92.98 & 33.48 & 54.90 & -9.43 &  63.57  &  -3.70  \\
        \textbf{ $\circ$ Africa} & 87.93 & 34.26 & 47.53 & -16.84 &  59.73  &  -7.54  \\
        \textbf{ $\circ$ East Asia} & 83.05 & 35.46 & 45.51 & -18.86 &  50.18  &  -17.09  \\
        \bottomrule
    \end{tabular}
    }
    \caption{Accuracy (\%) on the subset of each region in GD-VCR and the original VCR development set. With regard to Western regions, two models' performance gap of the original VCR development set and other regions is shown. We also report human's accuracy (\%) over each region subset. Annotators are from United Kingdom and United States according to MTurk. $^*$ denotes the reported result in~\citet{zellers2019vcr}.}
    \label{tab:overall_performance}
\end{table*}

\subsection{Model Performance}
\label{sec:model_performance}
We apply the two models on GD-VCR benchmark to study how well the two models can generalize. Results are shown in Table \ref{tab:overall_performance}. Key observations are summarized as follows:

\paragraph{Western vs. Non-Western Regions.} We find that the models perform significantly worse on the images from non-Western regions. According to VisualBERT's results, we observe that the gap between Western and South Asian regions is 9.43\%, while it greatly amplifies to 16.84\% and 18.86\% when it comes to the comparison with African and East Asian regions, respectively. These results reflect significant differences of models' reasoning ability on the examples from different regions. 

\paragraph{VisualBERT vs. ViLBERT.} We find that ViLBERT outperforms VisualBERT by 6.04\% on GD-VCR. We conjecture that the higher performance of ViLBERT partly results from the pre-training data: VisualBERT is pre-trained on COCO Captions which includes 80K images~\cite{Chen2015MicrosoftCC}, while ViLBERT's pre-training data are from Conceptual Captions containing 3.3M images~\cite{sharma-etal-2018-conceptual}. The larger coverage of image contents may help ViLBERT generalize to the images with regional characteristics. It is also shown that the performance gap over the images from Western and non-Western regions shrinks when applying ViLBERT. However, the gap is still significant, ranging from 3.70\% to 17.09\%.  

\paragraph{Western v.s. Original VCR Dataset.} We observe a performance gap around 2\%-6\% between images from Western and the original VCR dataset. We speculate that the gap is caused by one main aspect: the requirements in the image collection stage are slightly different. We expect to collect images containing regional characteristics, including cultural elements like customs. It may add to the complexity of the reasoning process as cultural commonsense is needed. However, the gap is much smaller compared with the gap between Western and other regions.

\subsection{Human Evaluation}
Apart from the model performance, we investigate how well human beings perform on GD-VCR. We randomly select 40 QA pairs of each region, and there are 160 QA pairs in total for evaluation. We recruit qualified annotators living in United Kingdom and United States from MTurk\footnote{The annotators should complete at least 1000 HITs, with an approval rate above 95\%.} to accomplish the evaluation. Assuming them to be familiar with Western culture, we are interested to see their performance on the examples from other regions.

Human evaluation results are shown in Table~\ref{tab:overall_performance}. We notice that human performance is much better than models. More importantly, we observe that the performance gap among regions is much smaller than that of two V\&L models. For example, annotators from Western can achieve 87.93\% accuracy on East Asian images, and the gap reduces to 8.18\% from 18.86\% and 17.09\%. It implies that human beings are more capable of applying their commonsense knowledge and transferring it to the comprehension in geo-diverse settings, while models are still far away from this level.

\section{Analyses of Performance Disparity}
As we observe large performance gaps between Western and non-Western data in Section~\ref{sec:model_performance}, in this section, we inspect the pre-trained V\&L model to analyze the reasons behind such performance disparity on two aspects, 1) \emph{regional differences of scenarios} and 2) \emph{reasoning level of QA pairs}. We analyze the \emph{VisualBERT} model, since its performance gap is more evident.

\subsection{Regional Differences of Scenarios}
As shown in Figure~\ref{fig:intro}, even the same scenarios such as wedding can take different visual forms across regions. Motivated by this, we investigate how large the performance gap is when we apply VisualBERT to the images of the same scenarios happening in different regions.

\begin{figure}[t]
    \centering
    \includegraphics[width=0.95\linewidth, trim=5 100 25 5, clip]{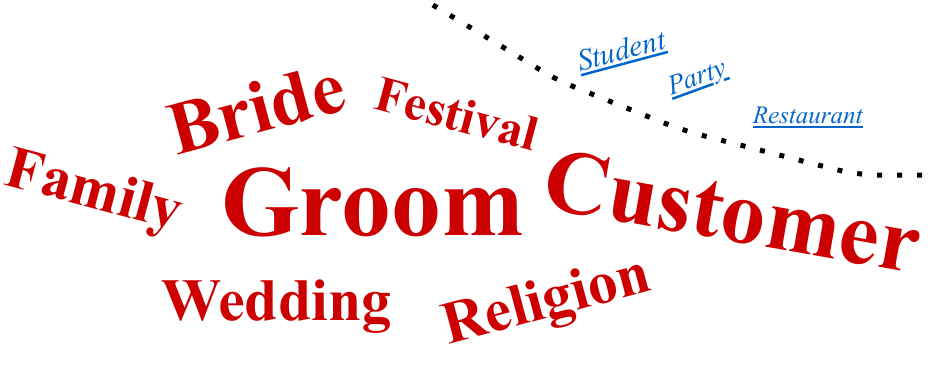}
    \caption{Visualization of the performance gap on images of the same scenarios in Western and non-Western regions. The larger the characters, the larger the performance gap over the scenarios. The \textbf{\textcolor[HTML]{CC0000}{red}} and \underline{\textbf{\textcolor[HTML]{0066CC}{blue}}} words are the scenarios whose performance gap is \emph{above} and \emph{below} 8\%, respectively. Detailed accuracy on these scenarios is shown in Appendix~\ref{appendix:detailed_accuracy}.}
    \label{fig:wordcloud}
\end{figure}

We select the scenarios that frequently appear in the annotated keyword set of GD-VCR. Specifically, we choose the scenarios which appear at least 10 times in \emph{not only} Western images, \emph{but also} the images from any of the other regions. We visualize each scenario's performance gap between the images from Western and non-Western regions in Figure \ref{fig:wordcloud}. The scenarios whose gap is above 8\% are colored in \textbf{\textcolor[HTML]{CC0000}{red}}; otherwise, they are labeled by \underline{\textbf{\textcolor[HTML]{0066CC}{blue}}}. 

\begin{table*}[t]
\centering
\scalebox{0.9}{
\begin{tabular}{lccccc}
\toprule
\multirow{2}{*}{\textbf{Regions}} & \multicolumn{2}{c}{\textbf{Low-order}} & \multicolumn{2}{c}{\textbf{High-order}} & \multirow{2}{*}{$|Low-High|$} \\
                         & \textbf{Acc. ($Low$)}       & \textbf{Gap (West)}      & \textbf{Acc. ($High$)}        & \textbf{Gap (West)}      &                        \\ \midrule
\textbf{West}                  & 65.15      & 0.00             & 66.60       & 0.00             & 1.45                   \\
\textbf{South Asia}               & 54.37      & -10.78           & 52.37       & -14.23           & 2.00                   \\
\textbf{East Asia}                & 58.74      & -6.41            & 50.47       & -16.13           & 8.27                   \\
\textbf{Africa}                   & 56.06      & -9.09            & 40.35       & -26.25           & 15.71   \\
\bottomrule
\end{tabular}
}
\caption{VisualBERT's accuracy (\%) on low-order and high-order cognitive QA pairs. ``Gap (West)'' denotes performance gap over the QA pairs of images from Western and non-Western regions. ``$|Low-High|$'' denotes the performance gap between low-order and high-order cognitive QA pairs from the same regions.}
    \label{tab:reasoning_level}
\end{table*}

As shown in Figure \ref{fig:wordcloud}, we find that on the scenarios which often contain regional characteristics (e.g., wedding, religion, festival), the performance gap is much larger, from 8.28\% to 23.69\%. 
One interesting finding is that, aside from festival, wedding and religion, which are generally considered to be different across regions, the gap is considerably large over the scenarios involving customers. 
We speculate that it is also due to regional characteristics. As shown in Figure \ref{fig:customer_example} of Appendix \ref{appendix:examples}, in East Asia and South Asia, many customers would buy things from the local merchants along the streets, while in Western regions, customers typically shop in supermarkets and restaurants. The visual discrepancy may result in errors on the ``customer'' scenarios in GD-VCR. 

On the other hand, for the scenarios such as party, restaurant and student, the gap is only 0.42\%, 1.29\% and 1.12\%, respectively. We notice that these scenarios are more similar across regions. For example, parties are related to actions like drinking, dancing, and celebration, which are common and take on similar visual forms across regions. Such similarity may contribute to model's high transfer performance on ``party''.

\subsection{Reasoning Level of QA Pairs}
The QA pairs in GD-VCR are \emph{high-order} cognitive QA pairs, which require several reasoning steps to be solved. For example, to infer that ``\textit{\colorbox[rgb]{ .891,  .891,  .891}{\textbf{[person1]}} and \colorbox[rgb]{ .891,  .891,  .891}{\textbf{[person2]}} are in a wedding}'' in the middle image of Figure~\ref{fig:intro}, human beings must first recognize basic facts such as \colorbox[rgb]{ .891,  .891,  .891}{\textbf{[person1]}} is wearing in red and her face is covered by a red cloth. Only by combining the recognized facts and regional commonsense can human make correct predictions.
Therefore, the model's failure on these high-order cognitive QA pairs from non-Western regions may be attributed to two reasons: 1) the model fails to recognize the basic facts from the image, 2) or the model succeeds on the basic facts but fails eventually due to lack of geo-specific commonsense.

To determine at which stage the model fails to generalize, we aim to answer the following two questions: \textbf{Q1.} \emph{Can the model perform similarly on recognizing basic visual information in the images from different regions?} \textbf{Q2.} \emph{Is the performance disparity attributed to the failure of understanding more advanced or basic visual information?}

According to the standard of reasoning level discrimination mentioned in Section~\ref{sec:question_design}, we categorize QA pairs into two types: \emph{low-order} and \emph{high-order} cognitive QA pairs. \emph{Low-order} cognitive QA pairs correspond to the inquiry on basic visual information, while \emph{high-order} QA pairs involve more advanced information.
Our analysis is mainly concerned with the two types of QA pairs.

\paragraph{Q1. Can model perform similarly on understanding basic visual information across regions?}
We evaluate model's performance on \emph{low-order} cognitive QA pairs to analyze this aspect.

As mentioned in Section~\ref{sec:question_design}, GD-VCR is composed of high-order cognitive QA pairs but without low-order pairs. Therefore, we additionally annotate low-order cognitive QA pairs on the images of GD-VCR. Specifically, we randomly select 30 images per region and design low-order QA pairs based on these selected images. Finally, we collect 22 QA pairs on Western images, 26 on East Asian images, 16 on South Asian images, and 22 on African images.

Results are shown in Table \ref{tab:reasoning_level}. We observe that the performance over the low-order cognitive QA pairs is all around 60\% for the four regions. Performance over Western images is still the highest among the four regions. But note that the performance gap between the images from Western and non-Western regions is not so large as the overall gap shown in Table~\ref{tab:overall_performance}. For example, the overall performance gap between East Asia and Western is around 19\%, but it decreases to 6.41\% when the model deals with simpler situations. It demonstrates that, when encountering the QA pairs focusing on simple recognition, VisualBERT can narrow down the gap on the images from different regions. In other words, VisualBERT shows more similar ability to process basic visual information, no matter where the images are from.

\paragraph{Q2. Is the performance disparity attributed to understanding on more advanced or basic visual information?}
We analyze the performance over \emph{low-order} and \emph{high-order} cognitive QA pairs. For a fair comparison, both types of QA pairs share \emph{the same images}.

Results are shown in Table \ref{tab:reasoning_level}. We observe that VisualBERT's performance over low-order cognitive QA pairs is higher than that over high-order QA pairs on images from East Asia, South Asia, and Africa. Especially, on the images from African regions, the performance gap between these two types of QA pairs is 15.71\%. 

Furthermore, from Table \ref{tab:reasoning_level}, we notice that the performance gap between Western and non-Western regions on high-order cognitive QA pairs is much larger than that on low-order QA pairs. For the images from East Asian regions, the performance gap with regard to Western regions on low-order pairs is 6.41\%. The gap amplifies to 16.13\% when VisualBERT is applied to high-order QA pairs. For African images, the gap changes rapidly from 9.09\% to 26.25\%. These results show that VisualBERT trained on VCR lacks the ability to perform complex reasoning on the scenarios in non-Western regions. We hope our findings could inspire future work to model high-level reasoning process better with geo-diverse commonsense knowledge in commonsense reasoning tasks.

\section{Conclusion}
We propose a new benchmark, GD-VCR, for evaluating V\&L models' reasoning ability on the QA pairs involving geo-diverse commonsense knowledge. Experimental results show that the V\&L models cannot generalize well to the images regarding the regional characteristics of non-Western regions. Based on VisualBERT's results, we find that 1) the scenarios such as wedding, religion and festival, which require geo-diverse commonsense knowledge to be understood, and 2) the reasoning difficulty of QA pairs are highly associated with the performance disparity. For broader impact, we hope that the GD-VCR benchmark could broaden researchers' vision on the scope of commonsense reasoning field and motivate researchers to build better commonsense reasoning systems with more inclusive consideration.

\section*{Acknowledgement}
We thank Xiao Liu, Ming Zhong, Te-Lin Wu, Masoud Monajatipoor, Nuan Wen, and other members of UCLANLP and UCLA PlusLab groups for their helpful comments. We also greatly appreciate the help of anonymous annotators for their effort into constructing the benchmark. This work was partially supported by DARPA MCS program under Cooperative Agreement N66001-19-2-4032. The views expressed are those of the authors and do not reflect the official policy or position of the Department of Defense or the U.S. Government.

\section*{Ethical Considerations}
\label{ethics}
In this work, we propose a geo-diverse visual commonsense reasoning dataset GD-VCR. Since the paper introduces new dataset, we discuss the potential ethical issues about data collection.

\paragraph{Intellectual Property and Privacy Rights.}
We ensure that intellectual property and privacy rights of the original authors of videos and recruited annotators are respected during the dataset construction process with permission of licence\footnote{Fair use on YouTube. \url{support.google.com/youtube/answer/9783148?hl=en}}\footnote{Copyright Law of the People's Republic of China (Article 22). \url{http://www.gov.cn/flfg/2010-02/26/content_1544458.htm}.}. We also claim that the collected data would not be used commercially.

\paragraph{Compensation for Annotators.}
We recruit annotators from Amazon Mechanical Turk platform\footnote{\url{www.mturk.com}} and college departments of foreign languages and culture. In image collection stage, we paid annotators \$0.5-0.7 per collected image. In QA pair annotation stage, the payment is \$0.2 per QA pair. For validation and human evaluation, we pay them \$0.02-0.03 per QA pair. The pay rate is determined by a preliminary annotation trial to ensure the average hourly pay rate is around \$12 per hour. 
The annotations on the images from East Asian regions are partly done by the authors of this work.

\paragraph{Potential Problems.}
Although we have considered the potential geographic bias in the benchmark construction process, GD-VCR may still contain unwanted bias. First, due to the resource constraints, GD-VCR dataset is unable to cover diverse regional characteristics at once. For instance, we do not take Southeast Asian, Arabic and Latin American regions into account. Moreover, even groups in the same region may have different beliefs. For the regions like Africa, the regional differences between West Africa, East Africa, and North Africa are evident. However, in GD-VCR, the images from Africa are mainly sourced from East Africa. It inevitably introduces geographic bias into our benchmark. More fine-grained analysis should be conducted to scale up this study, especially before the visual commonsense reasoning model is used in the commercial product.

\bibliography{nlp,ref}
\bibliographystyle{acl_natbib}

\appendix

\clearpage

\section{Additional Details of Annotation Pipeline}
\label{sec:appendix}
\subsection{Image Collection}
\label{appendix:image_collection_1}
In addition to the requirements mentioned in Section~\ref{sec:image_collection}, we have additional requirements on the contents, quality, and sources of images. The image should have at least two people, and should not be grainy and blurry. We require annotators to choose movies, TV series, documentaries and trailers which are free to access and do not have copyright issues. Together with the images and their keywords, we also collect video names, screenshot timestamps, and the links of videos. It is to help the annotators in later stages better understand the image contents with video contexts.

\begin{table*}[t]
\centering
\scalebox{0.8}{
\begin{tabular}{lcccccccccc}
\toprule
\textbf{Regions}    & \textbf{Wedding} & \textbf{Festival} & \textbf{Religion} & \textbf{Bride} & \textbf{Groom} & \textbf{Restaurant} & \textbf{Family} & \textbf{Student} & \textbf{Customer} & \textbf{Party} \\ \midrule
\textbf{West}    &   62.22      &   68.89   &    58.33      &  66.67     &   69.14    &   61.90         &   59.26     &   61.54      &   66.67    &   55.83   \\
\textbf{Other Regions} &   50.00      &   60.61       &    46.21      &  52.78     &   45.45    &    60.61        &   47.27     &   60.42      &  44.44   &    56.25     \\ \bottomrule
\end{tabular}
}
\caption{Accuracy (\%) on the images involving the same scenarios from different regions.}
    \label{tab:accuracy}
\end{table*}

\begin{table}[ht]
\centering
\scalebox{0.8}{
\begin{tabular}{l}
\toprule
\textbf{Question Templates}  \\
\midrule
\small 1. What did sb. (just) v.? \\
\small 2. What did sb do when/before/as \texttt{CLAUSE}? \\
\small 3. What (n.) is sb. v.+ing prep. \texttt{PHRASE}? \\
\small 4. What (n.) is sb. v.+ing? \\
\small 5. What is sb.'s job/occupation? \\ 
\small 6. What is sb.'s relationship with sb.? \\
\small 7. Why is sb. v.+ing sth. \texttt{CLAUSE}? \\
\small 8. Why is sb. adj.? \\
\small 9. Why is sb. here? \\
\small 10. Why is sb. acting adv.? \\
\small 11. How does sb. feel/look? \\
\small 12. Where are sb. (v.+ing)? \\
\small 13. What will sb v. next/is about to do? \\
\small 14. What will sb v. when/after \texttt{CLAUSE}? \\
\small 15. What will sb v. (if) \texttt{CLAUSE}? \\
\small 16. Where will sb. go? \\
\small 17. Where was sb. previously? \\ \bottomrule
\end{tabular}
}
\caption{Question template list summarized from the original VCR development set.}
\label{tab:template_list}
\end{table}

\subsection{QA Pair Annotation}
\label{appendix:qa_pair_annotation}
\paragraph{Question Template List.} As mentioned in Section~\ref{sec:qa_annotation}, we recommend annotators to design questions based on question templates. The template list is shown in Table \ref{tab:template_list}. 
For clustering methods to summarize templates, we use K-Means algorithm to cluster similar question patterns. Specifically, the maximum number of clusters is at most 20 clusters. The algorithm will automatically stop until the 200-th iteration. 

\paragraph{Other Annotation Details.} To pursue diversity of question types, we require annotators to design questions via different question templates. 
Besides, we ask annotators to avoid annotating too simple answers, such as ``\textit{yes}'' and ``\textit{no}''.

\section{Details of Answer Candidate Generation Algorithm}
\subsection{Relevance Model}
\label{appendix:relevance}
Relevance model is to evaluate the relevance score between questions and answers. Higher relevance scores indicate that the answers are more relevant with the questions. We train the relevance model based on pre-trained BERT-base parameters~\cite{wolf-etal-2020-transformers}. Specifically, the training data are all from the \emph{original VCR training set} and composed by relevant and irrelevant QA pairs. The relevant QA pairs are the ones consisting of questions and their corresponding right answers; the irrelevant pairs are the ones consisting of questions and random answers sampled from the whole set of answer choices. We build a binary classifier upon these training data to classify whether an answer is relevant with a question or not. The relevance score is the probability of being relevant pairs.

\subsection{Pseudo Code of Answer Candidate Generation Algorithm}
\begin{algorithm}[H]
    \small
    \caption{Answer Candidate Generation Alg.}
    \textbf{Input:} Question $Q = \{q_1, q_2, ..., q_n\}$, the question's right answer $Corr = \{c_1, c_2, ..., c_n\}$, answer candidate pool $\mathbf{A} = \{A_1, A_2, ..., A_m\}$, relevance model $\mathrm{Rel}$, similarity model $\mathrm{Sim}$. $q_i$ and $c_i$ indicate tokens. \\
    \textbf{Output:} The whole set of four answer choices $\mathbf{ansList}$ of the given question $Q$, including one right choice $Corr$ and three answer candidates $W_1$, $W_2$, and $W_3$. 
    \begin{algorithmic}[1]
     \State Initialization: $\mathbf{ansList}$ $\gets$ $\{Corr\}$.
     \For{$t=1,2,3$}
        \For{each $A_i$ in $\mathbf{A}_{\lfloor(t-1)\times\frac{m}{3}\rfloor+1:\lfloor t\times\frac{m}{3}\rfloor}$}
            \State Initialization: $score$ $\gets$ $0$, $minscore$ $\gets$ $+\infty$.
            \If {$\mathrm{Rel}(Q,A_i)\geq0.9$}
                \For{each ${ansList}_i$ in $\mathbf{ansList}$}
                    \State $similarity \gets$ $\mathrm{Sim}({ansList}_i,A_i)$
                    \If {$similarity\leq0.2$}
                        \State $score \gets$ $score+similarity$
                    \Else
                        \State $score \gets$ $score+10$
                    \EndIf
                \EndFor
                \If {$score \textless minscore$}
                    \State $minscore \gets score$
                    \State $W_t \gets A_i$
                \EndIf
            \EndIf
        \EndFor
        \State $\mathbf{ansList}$ $\gets$ $\mathbf{ansList} \cup \{W_t\}$
     \EndFor \\
     \Return $\mathbf{ansList}$
    \end{algorithmic}
    \label{alg:wrong_choice_generation}
\end{algorithm}
\label{appendix:wrong_choice}
The pseudo code of the algorithm is shown in Algorithm \ref{alg:wrong_choice_generation}. The two principles for selecting answer candidates are as follows: for each QA pair, 1) they should be relevant with the questions; 2) they should not be similar with the right choices and the selected answer candidates. The model that computes similarity is the ``stsb-roberta-base'' model~\cite{reimers-2019-sentence-bert} from \url{github.com/UKPLab/sentence-transformers}.

\section{Implementation Details of Fine-tuning VisualBERT and ViLBERT}
\label{appendix:implementation}
Following VisualBERT (135.07M parameters) configuration on VCR\footnote{\url{github.com/uclanlp/visualbert}}, we directly use the model pre-trained on COCO~\cite{Chen2015MicrosoftCC} and original VCR training set. The experiments of ViLBERT\footnote{\url{github.com/jiasenlu/vilbert_beta}} (252.15M parameters) is based on the model pre-trained on Conceptual Captions~\cite{sharma-etal-2018-conceptual} and original VCR training set. Both models are then fine-tuned for 8 epochs on 4 NVIDIA GeForce GTX 1080 Ti GPUs, with learning rate $2e-5$. The batch size of VisualBERT and ViLBERT is 32 and 16, and fine-tuning one epoch with VisualBERT and ViLBERT costs 5.28 and 6.75 hours, respectively. For both models, we choose the epoch which performs the best on the original VCR development set among 8 epochs.

\section{Accuracy on the QA Pairs Involving Specific Scenarios}
\label{appendix:detailed_accuracy}
Table~\ref{tab:accuracy} shows VisualBERT's accuracy of the QA pairs involving specific scenarios depicted in Figure~\ref{fig:wordcloud}. Besides the study on GD-VCR, we also make comparison between model performance on GD-VCR and the original VCR development set. We select the scenarios frequently appearing in both GD-VCR and the original VCR development set. 

Results are shown in Table~\ref{tab:compare_with_vcr}. We observe that on the images involving scenarios such as funeral, VisualBERT's performance gap is nearly 25\%, which is considerably large. The results further demonstrate that the model is still incapable of tackling the QA pairs which are involving cultural differences behind scenarios well. 

\begin{figure}[t]
    \centering
    \includegraphics[width=\linewidth, trim=10 55 60 70, clip]{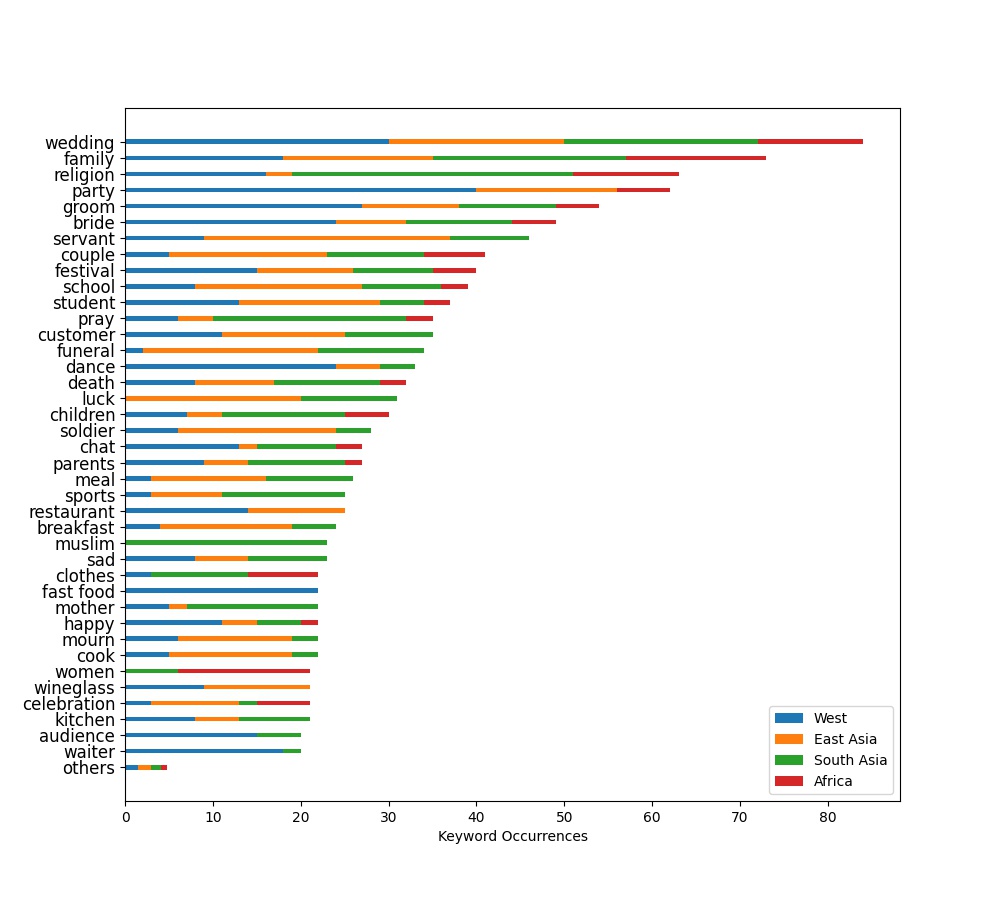}
    \caption{Statistics of keyword occurrences. ``Others'' denotes the average occurrences of the keywords appearing less than 20 times.}
    \label{fig:keyword_occurrences}
\end{figure}

\begin{table}[t]
\centering
\scalebox{0.8}{
\begin{tabular}{lcccc}
\toprule
\textbf{Regions/Datasets}    & \textbf{Wedding} & \textbf{Funeral} & \textbf{Servant} \\ \midrule
\textbf{Original VCR}    &   71.78      &   55.00   &    59.72    \\
\textbf{Other Regions} &   50.00      &    30.25      &    48.81      \\ \bottomrule
\end{tabular}
}
\caption{Accuracy (\%) on the images involving the same scenarios from the original VCR dataset and non-Western regions from GD-VCR dataset, respectively.}
    \label{tab:compare_with_vcr}
\end{table}

\section{Keywords in GD-VCR Dataset}
Figure \ref{fig:keyword_occurrences} shows the overall statistics of keyword occurrences in GD-VCR benchmark. There are 693 keywords in total, showing the diversity of the scenarios covered by GD-VCR dataset. Besides, we observe that the keywords whose corresponding scenarios have evident regional differences, such as ``\textit{wedding}'', ``\textit{religion}'', ``\textit{groom}'', ``\textit{bride}'', appear frequently in GD-VCR.  

\section{More Examples in GD-VCR Dataset}
\label{appendix:examples}
\begin{figure*}[t]
    \centering
    \includegraphics[width=\linewidth, trim=0 0 100 0, clip]{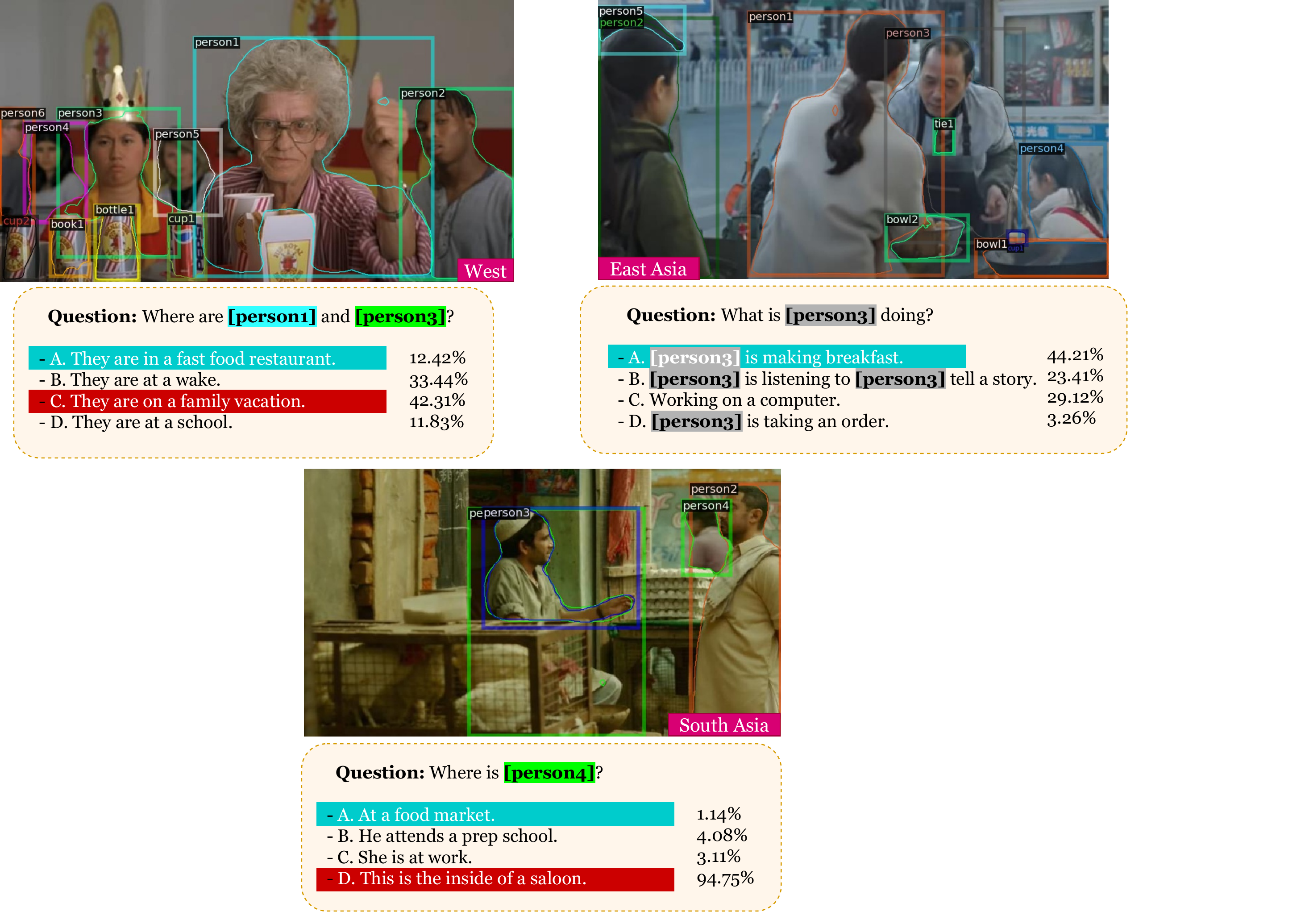}
    \caption{Examples of the images regarding ``\textit{customer}''. Left-to-right order: Western, South Asia, East Asia. We visualize the prediction of the VisualBERT model fine-tuned on the original VCR training set. The blue blocks denote the right answer choices. If red block appears, it means that VisualBERT wrongly predict the answer. The rightmost value indicates the probability of the corresponding choices being selected by VisualBERT.}
    \label{fig:customer_example}
\end{figure*}

\begin{figure*}[t]
    \centering
    \includegraphics[width=\linewidth, trim=5 155 335 0, clip]{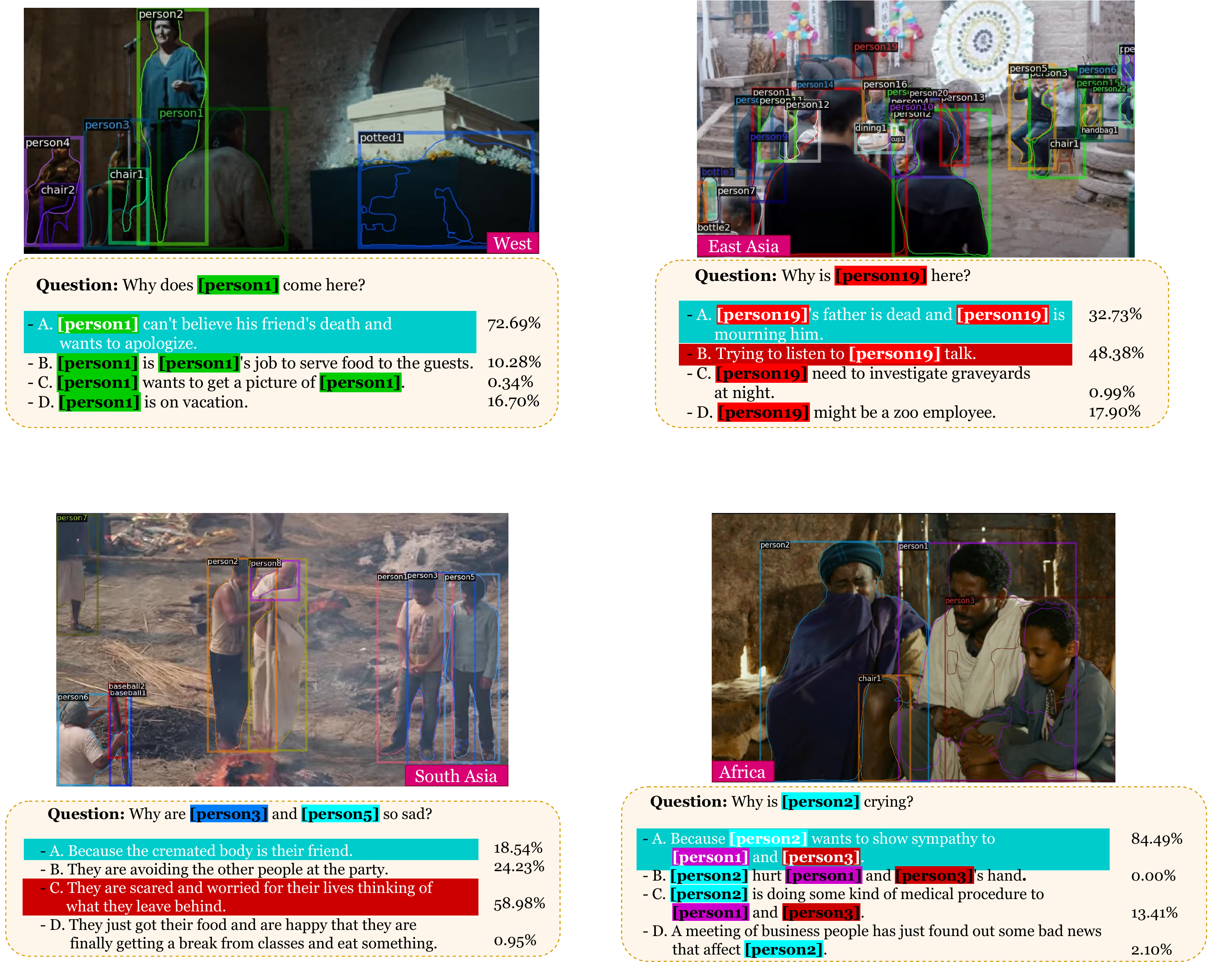}
    \caption{Examples of the images regarding ``\textit{funeral}'' or ``\textit{death}''. Left-to-right order in the first row: Western, East Asia; Left-to-right order in the second row: South Asia, Africa.}
    \label{fig:funeral_example}
\end{figure*}

In this section, we showcase several examples of GD-VCR in detail. Aside from the images about ``\textit{wedding}'' in Figure \ref{fig:intro}, we manifest the images regarding to ``\textit{customer}'' and ``\textit{funeral}'' from the four regions we study. In Figure \ref{fig:customer_example} and Figure \ref{fig:funeral_example}, we can observe the regional characteristics from the selected images. Furthermore, we visualize VisualBERT's prediction on each QA pair. 

\end{document}